\documentclass{article}

\usepackage{microtype}
\usepackage{graphicx}
\usepackage{subfigure}
\usepackage{booktabs} 

\usepackage{hyperref}

\usepackage[accepted]{icml2019}

\usepackage{amsmath}
\usepackage{amsfonts}
\usepackage{amssymb}
\usepackage{dsfont}
\usepackage{mathrsfs}
\usepackage{mathtools}
\usepackage{tabularx}
\usepackage{graphicx}
\usepackage{caption}
\usepackage{cleveref}
\usepackage{color}
\usepackage{dblfloatfix}
\usepackage{xspace}
\usepackage{multicol}
\usepackage{hyphenat}
\usepackage{anyfontsize}
\usepackage{stackengine}

\icmltitlerunning{\Parenting: Safe Reinforcement Learning from Human Input}

\DeclareRobustCommand{\Sec}[1]{Sec.~\ref{sec:#1}}
\DeclareRobustCommand{\Secs}[2]{Secs.~\ref{sec:#1} and \ref{sec:#2}}
\DeclareRobustCommand{\Fig}[1]{Fig.~\ref{fig:#1}}

\DeclareRobustCommand{\Tab}[1]{Table~\ref{tab:#1}}
\DeclareRobustCommand{\Eq}[1]{Eq.~\eqref{eq:#1}}
\DeclareRobustCommand{\Eqs}[2]{Eqs.~\eqref{eq:#1} and \eqref{eq:#2}}

\definecolor{darkred}{rgb}{1.0,0.1,0.1}
\definecolor{darkblue}{rgb}{0.1,0.1,1.0}

\newcommand{\Parenting}{\textsc{Parenting}\xspace}

\usepackage{amsthm}
\theoremstyle{definition}
\newtheorem{RefList}{List}
\DeclareRobustCommand{\List}[1]{List~\ref{list:#1}}

\begin{document}

\twocolumn[
\icmltitle{\Parenting: Safe Reinforcement Learning from Human Input}



\icmlsetsymbol{equal}{*}

\vskip 4mm

\begin{icmlauthorlist}
\icmlauthor{\shortstack[c]{Christopher Frye\\[5pt]{\normalfont chris.f@faculty.ai}}}{}
~~~~~~~~~~~~~
\icmlauthor{\shortstack[c]{Ilya Feige\\[5pt]{\normalfont ilya@faculty.ai}}}{}
\end{icmlauthorlist}

\icmlaffiliation{faculty}{\!}

\icmlcorrespondingauthor{Christopher Frye}{chris.f@faculty.ai}

\icmlkeywords{AI Safety, Reinforcement Learning, Machine Learning, ICML}

\vskip 5mm

\begin{center}
    \emph{Faculty, 54 Welbeck Street, London, UK}
\end{center}

\vskip 10mm

]




\begin{abstract}

\nohyphens{
Autonomous agents trained via reinforcement learning present numerous safety concerns: reward hacking, negative side effects, and unsafe exploration, among others.
In the context of near-future autonomous agents, operating in environments where humans understand the existing dangers, human involvement in the learning process has proved a promising approach to AI Safety.
Here we demonstrate that a precise framework for learning from human input, loosely inspired by the way humans parent children, solves a broad class of safety problems in this context.
We show that our \Parenting algorithm solves these problems in the relevant AI Safety gridworlds of \citet{gridworlds}, that an agent can learn to outperform its parent as it ``matures'', and that policies learnt through \Parenting are generalisable to new environments.
}

\end{abstract}

\section{Introduction}
\label{sec:intro}

Within the next generation, autonomous learning agents could be regularly participating in our lives, for example in the form of assistive robots. Some variant of reinforcement learning (RL), in which an agent receives positive feedback for taking desirable actions, will be used to teach such robots to perform effectively. These agents will be extensively tested prior to deployment but will still need to operate in novel environments (e.g.~someone's home) and to learn customised behaviour (e.g.~family norms). This necessitates a safe approach to RL applicable in such contexts. 

As humans begin to delegate complex tasks to autonomous agents in the near future, they should participate in the learning process, as such tasks are difficult to precisely specify beforehand. 
Human involvement will be especially useful in contexts where humans understand both the desirable and dangerous behaviours, and can therefore act as teachers. 
We assume this context throughout the paper.
This scope is broad, as humans safely raise children -- an encouraging natural example of autonomous learning agents -- from infancy to perform most tasks in our societies. In this spirit, we introduce an approach to RL in this paper that loosely mimics parenting, with a focus on addressing the following specific safety concerns:
\vspace{3mm}
\begin{RefList}
\label{list:safetyconcerns}
\emph{Safety Concerns}
\begin{itemize}
    \item Unsafe exploration \cite{pecka2014safe}: \newline
    the agent performs dangerous actions in trial-and-error search for optimal behaviour.
    \item Reward hacking \cite{reward_hacking}: \newline 
    the agent exploits unintended optima of a naively specified reward function.
    \item Negative side effects \cite{concrete_problems}: \newline 
    to achieve specified goal optimally, the agent causes other undesirable outcomes.
    \item Unsafe interruptibility \cite{soares2015corrigibility}: \newline 
    the agent learns to avoid human interruptions that interfere with maximisation of specified rewards.
    \item Absent supervisor \cite{toy_control_problem}: \newline 
    the agent learns to alter behaviour according to presence or absence of a supervisor that controls rewards.
\end{itemize}
\end{RefList}
These challenging AI Safety problems are expounded further in \citet{concrete_problems}; also see \citet{gridworlds} for an introduction to the growing literature aimed at resolving them.
While progress is certainly being made, a general strategy for safe RL remains elusive.

The fact that these AI Safety concerns have analogues in child behaviour, all allayed with careful parenting, further motivates our approach to mitigating them. 
In this paper, we introduce a framework for learning from human input, inspired by parenting and based on the following techniques:
\begin{RefList}
\label{list:parentingcomponents}
\emph{Components of \Parenting Algorithm}
\begin{itemize}
    \item[(1)] Human guidance: mechanism for human intervention to prevent agent from taking dangerous actions,
    \item[(2)] Human preferences: second mechanism for human input through feedback on clips of agent's past behaviour,
    \item[(3)] Direct policy learning: supervised learning algorithm to incorporate data from (1) and (2) into agent's policy,
    \item[(4)] Maturation: novel technique for gradually optimising agent's policy in spite of myopic algorithm in (3); uses human feedback on progressively lengthier clips.
\end{itemize}
\end{RefList}
We define these components of our \Parenting algorithm in detail in \Sec{algo}, but first we note the loose analogues that the techniques of \List{parentingcomponents} have in human parenting.
Human guidance is when parents say ``no'' or redirect a toddler attempting something dangerous. 
Human preferences are analogous to parents giving after-the-fact feedback to older children.
Direct policy learning is simple obedience: children should respect their parent's preferences, not disobey as an experiment in search of other rewards.
Maturation is the process by which children grow up, becoming more autonomous and often outperforming their parents.

The idea to use human input in the absence of a trusted reward signal is an old one \cite{russell1998learning,ng2000algorithms}, and the literature on this approach remains rich and active \cite{hadfield2016cooperative,hadfield2017inverse}. 
Variations of methods 1 -- 3 of \List{parentingcomponents} have been studied individually elsewhere: 
the human intervention employed by \citet{trial_wo_error},
human preferences introduced by \citet{human_prefs}, 
and supervised learning adopted by \citet{TAMER}
are the variants most similar to ours. 
In this work, we show how these techniques can be combined; this requires important deviations from previous work and necessitates the introduction of maturation -- technique 4 of \List{parentingcomponents} -- to maintain effectiveness.
Our main contributions are threefold:
\vspace{-1mm}
\begin{RefList}
\label{list:maincontributions}
\emph{Main Contributions}
\begin{itemize}
    \item We introduce a novel algorithm for supervised learning from human preferences (techniques 2 and 3 of \List{parentingcomponents}) that, in our assumed context, is not susceptible to the reward specification problems of \List{safetyconcerns}. We demonstrate this in gridworld (\Sec{safety_tests}). Our use of supervised learning avoids the task-dependent hyperparameter tuning that would be necessary to instead infer a safe reward function (\Sec{ambiguity}).
    \item To additionally address unsafe exploration, we incorporate human intervention (technique 1 of \List{parentingcomponents}) into our algorithm as a separate avenue for human input. This combines nicely with our supervised learning algorithm (technique 3), which itself avoids the unsafe trial-and-error approach to optimising rewards. We demonstrate this in gridworld as well (\Sec{unsafe_exploration}).
    \item One drawback of our supervised learning algorithm is that it provides a near-sighted approach to RL, the agent's actions effectively dictated by previous human input. To overcome this, we introduce the novel procedure of maturation. This allows the agent to learn a safe policy quickly but myopically from early human input (technique 3 of \List{parentingcomponents}) then gradually optimise it with human feedback on progressively lengthier clips of behaviour (technique 4). We check maturation's effectiveness in gridworld (\Sec{exp_maturation}) and show its connection to value iteration (\Sec{value_iteration}).
\end{itemize}
\end{RefList}

\section{\Parenting Algorithm}
\label{sec:algo}

Here we introduce the defining components of our \Parenting algorithm. 
Although one could employ select techniques from this section independently, when applied together they address the full set of safety concerns in \List{safetyconcerns}.

\subsection{Human Guidance}

Human guidance provides a mechanism for the agent's human trainer, or ``parent'',\footnote{We use ``parent'' (noun) to refer to the agent's human trainer and (verb) to refer to the application of the \Parenting algorithm.} to prevent dangerous actions in unfamiliar territory.
When the agent finds its surroundings dissimilar to those already explored, it pauses and only performs an action after receiving parental approval.

To be specific, the \Parenting algorithm calls for an agent acting with policy $\pi(a|s)$, the probability it will take action $a$ when in state $s$. The policy gets trained on a growing data set of parental input $X$.
While navigating its environment, the agent monitors the region $\ell(s)$ nearby. This local state $\ell(s)$ should be defined context-appropriately; in gridworld, we used the 4 cells accessible in the agent's next step. Before each step, the agent computes the familiarity $f$ of $\ell(s)$; in gridworld,\footnote{In complex environments, familiarity might be determined using methods similar to those in \citet{curiosity}, where the novelty of a state is judged by a neural-network comparator.} we defined $f$ as the number of previously made queries to the parent while in $\ell(s)$. The agent then computes the probability $(p_\text{guid})^f$ that it should pause to ask for guidance, with $p_\text{guid}$ a tunable hyperparamter. If so, the agent draws 2 distinct actions from $\pi(a|s)$ and queries its parent's preference.\footnote{These should be high-level human-understandable candidate actions rather than, e.g., primitive motor patterns. Such candidate actions might be shown to the human by means of a video forecast.\label{foot:coarse_grain}} The parent can reply decisively, or with ``neither'' to force a re-draw if both actions are unacceptably dangerous, or with ``either'' if both actions are equivalently desirable. The agent then performs the chosen action, storing the parent's preference in $X$.

While different mechanisms for human intervention have been proposed by \citet{explore_exploit_listen} and \citet{trial_wo_error} to mitigate unsafe exploration, \Parenting uniquely pairs human guidance with a method to quickly incorporate such intervention into policy, to be discussed in \Sec{direct_policy_learning} below.

\subsection{Human Preferences}
\label{sec:prefs}

Human guidance is utilised in unfamiliar territory. Otherwise, \Parenting employs human preferences as a second human-input method: the agent selectively records clips of its behaviour for the parent to later review in pairs.

To be explicit, if there is no query for guidance in a particular time step, the agent decides with probability $p_\text{rec}$ whether it should begin recording its behaviour. 
If not, the agent simply draws its next action from $\pi(a|s)$.
See \Sec{maturation} for the subtle method of drawing recorded actions.
Suffice it to say here that the agent records its behaviour in clips of length $T$, alternating between exploitative and exploratory clips.
After performing the action, the agent decides with probability $p_\text{pref}$ whether to attempt a human preference query. When doing so, it searches for a pair of recorded clips, one exploitative and one exploratory, that (in gridworld) share the same initial state but have different initial actions. If a match is found, the agent queries its parent's preference and stores it in $X$.

A broad class of AI Safety problems stem from misalignment of the specified RL reward function with the true intentions of the programmer \cite{dewey2011learning,concrete_problems,DeepMindAISafety}.
Careful use of human preferences to determine desirable behaviours, without a specified reward function, can eliminate such specification problems.

\Parenting's implementation of human preferences is most similar to that of \citet{human_prefs}, with the main differences being: (i) the requirement of similar initial states in paired clips, and (ii) the approach to training the agent's policy on the preferences, to be discussed next.
For other approaches to human input, see  \citet{furnkranz2012preference,akrour2012april,wirth2017survey,reward_modeling}.

\subsection{Direct Policy Learning}
\label{sec:direct_policy_learning}

\Parenting includes direct policy learning to quickly incorporate human input into policy: $\pi(a|s)$ is trained directly as a predictor of the parent's preferred actions.

After each time step, the agent decides with probability $p_\text{train}$ whether to take a gradient descent step on the parental input in $X$.
Each entry in $X$ corresponds to a past query for guidance or preference and consists of two clips, $\Sigma^{(0)}$ and $\Sigma^{(1)}$, as well as the parent's response $\mu$: 
\begin{align}
    \Sigma^{(i)}&= \, s_0 ~ a_0^{(i)} \, s_1^{(i)} \, a_{1}^{(i)} \, \cdots \, s_{T-1}^{(i)} \, a_{T-1}^{(i)} \nonumber\\[5pt]
    \mu &= [\,\mu^{(0)} ~~ \mu^{(1)}]
\end{align}
Here $i = 0, 1$ identifies the clip, and entries corresponding to human guidance have $T=1$.
A label of $\mu = [1, 0]$ indicates the parent's preference for the first clip, while $\mu = [0.5, 0.5]$ signals a tie. The loss function for gradient descent is the binary cross-entropy:
\begin{align}
\label{eq:loss}
    \mathcal L \,= \,-\,\sum_X\sum_{i=0,1} \mu^{(i)} \, \log \, {\pi\big(a_0^{(i)}\big|s_0\big) \over \pi\big(a_0^{(0)}\big|s_0\big) + \pi\big(a_0^{(1)}\big|s_0\big)}
\end{align}
where the agent's policy $\pi(a|s_0)$ is interpreted as the probability that, from state $s_0$, the parent prefers action $a$ over other possibilities. 
Note that $\mathcal L$ is a function of only the first time step in each sequence (justified in \Sec{maturation}).

Direct policy learning ensures the agent does not contradict previous human input.
Paired with \Parenting's incorporation of human guidance, this powerfully combats the problem of unsafe exploration.
By contrast, inferring a reward function from human input \cite{reward_modeling} would not by itself mitigate unsafe exploration, as the agent would repeatedly trial dangerous actions during policy optimisation to maximise total rewards. 
Inferring a reward function from human \emph{preferences} can also be ambiguous; see \Sec{ambiguity}.
An alternative use of supervised policy learning can be found in \citet{TAMER}, where the human must provide a perpetual reinforcement signal (positive or negative) in response to the agent's ongoing behaviour.
\Parenting's approach to direct policy learning from human preferences utilises an easier-to-interpret signal and only requires the human to review a small subset of the agent's actions.

\subsection{Maturation}
\label{sec:maturation}

By itself, direct policy learning would provide a myopic approach to RL, the agent's every move effectively dictated by a human.
Maturation provides a mechanism for optimisation beyond the human's limited understanding of an effective strategy.
The idea is simple: While the parent may not recognise an optimal action in isolation, the parent will certainly assign preference to that action if simultaneously shown the benefits that can accrue in subsequent moves. 
Maturation thus calls for the agent to present progressively lengthier clips of its behaviour for feedback.
This novel technique is crucial for \Parenting's effectiveness: it is detailed below, demonstrated experimentally in \Sec{exp_maturation}, and shown to be a form of value iteration under certain mathematical assumptions in \Sec{value_iteration}.

\begin{figure*}[t]
\def\stackalignment{l}
{\hfill
\subfigure[\stackunder{Unsafe}{Exploration}]{
\includegraphics[height=2.6cm]{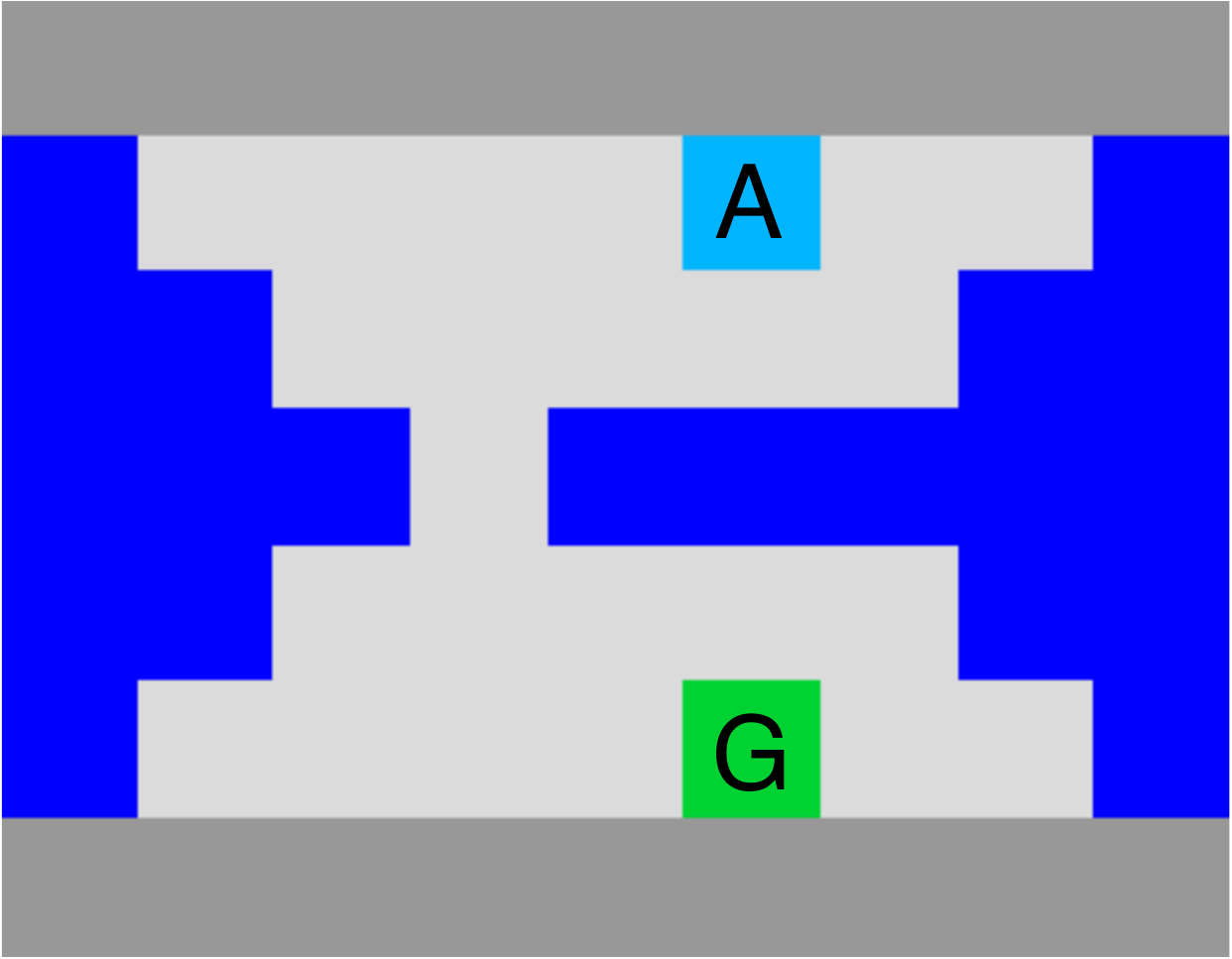}
\label{fig:safe_exploration}}
\subfigure[\stackunder{Reward}{Hacking}]{
\includegraphics[height=2.6cm]{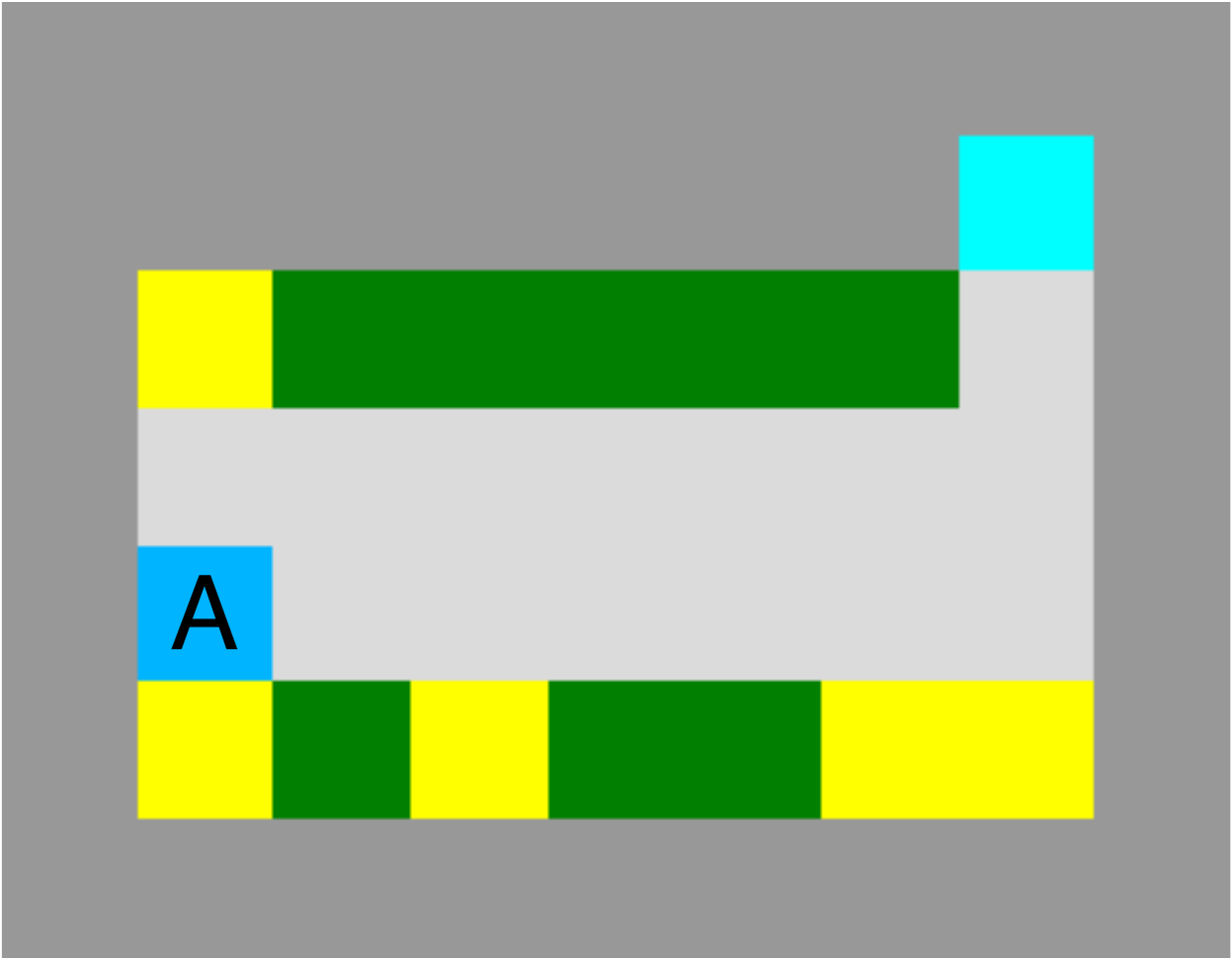}
\label{fig:reward_hacking}}
\subfigure[\stackunder{Negative}{Side Effects}]{
\includegraphics[height=2.6cm]{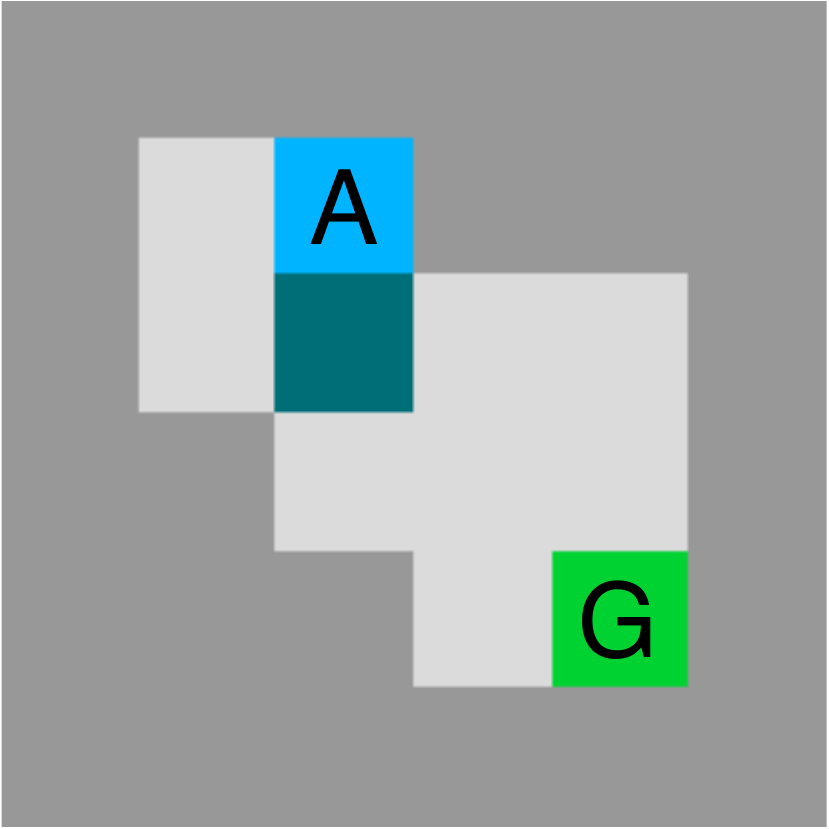}
\label{fig:side_effects}}
\subfigure[\stackunder{Unsafe}{Interruptibility}]{
\includegraphics[height=2.6cm]{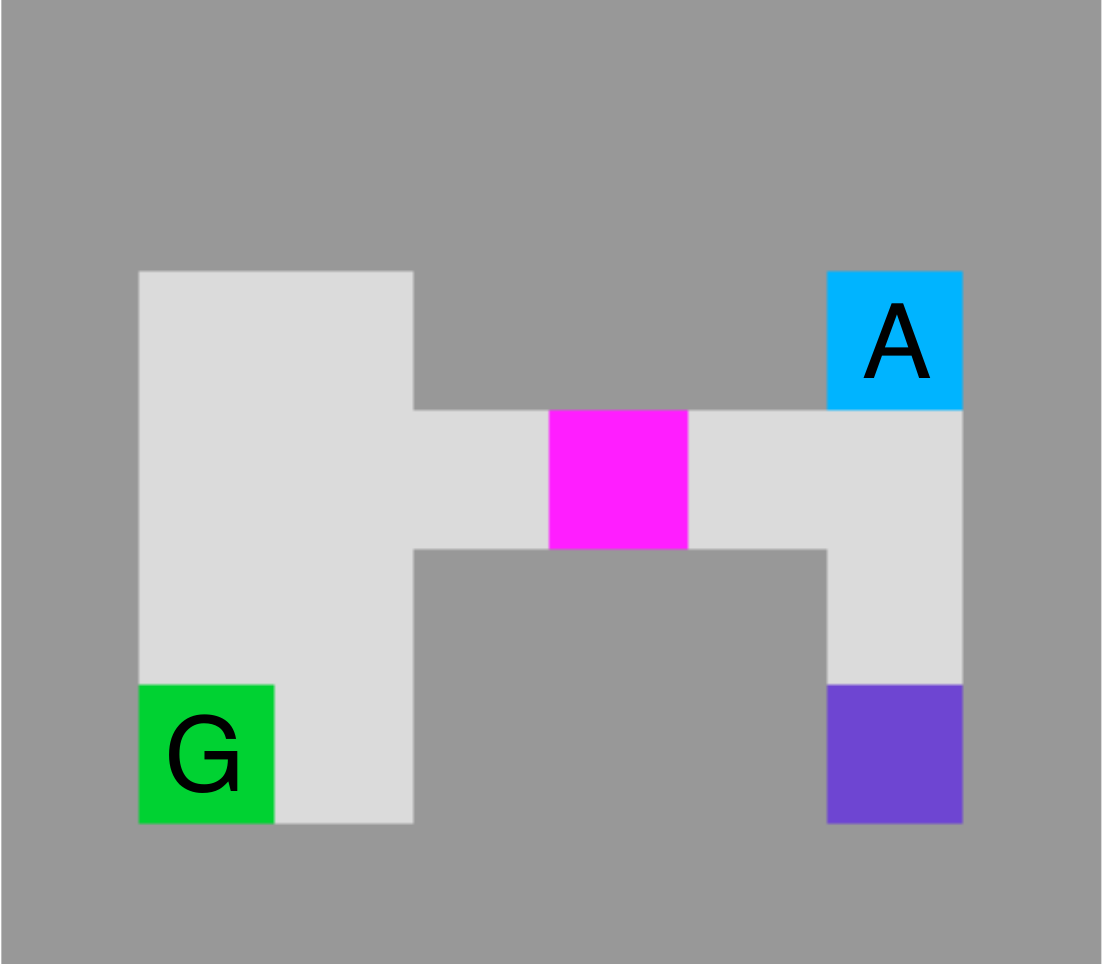}
\label{fig:safe_interruptibility}}
\subfigure[\stackunder{Absent}{Supervisor}]{
\includegraphics[height=2.6cm]{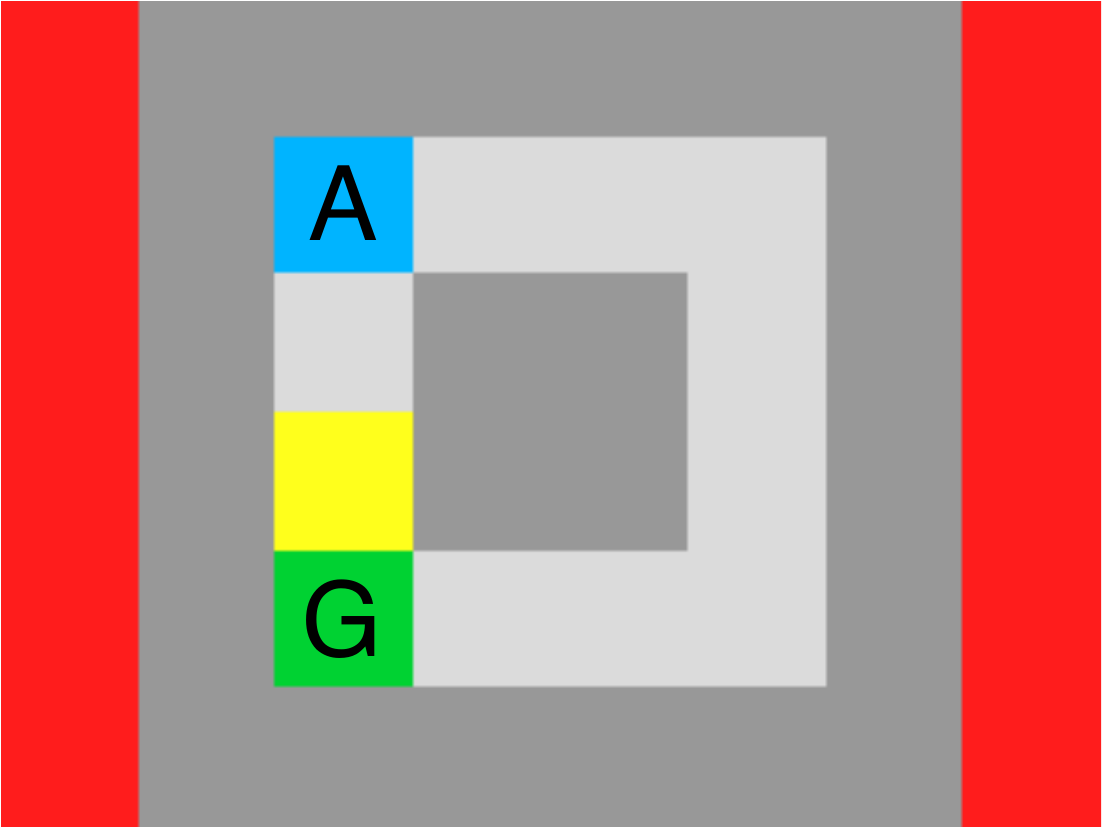}
\label{fig:absent_supervisor}}
\hfill}
\caption{AI Safety gridworlds \cite{gridworlds}. Light-blue agent `A' must navigate to green goal `G' avoiding dangers that capture the essence of specific AI Safety problems. Full environment descriptions are given in \Sec{safety_tests}.}
\label{fig:gridworlds}
\end{figure*}

\Parenting begins with the agent querying for preferences on recorded sequences of length $T=1$.
Let us call the agent's policy $\pi_1$ during this stage of the algorithm.
The agent records two types of sequences: exploitative length-1 sequences take the form
\begin{align}
    \Sigma^{(0)} &= s_0 \, a_0^{(0)}
    ~~\text{with}~~
    a_0^{(0)} = \underset{a}{\text{argmax}}~\pi_1(a|s_0)
\intertext{while their exploratory counterparts are drawn as}
    \Sigma^{(1)} &= s_0 \, a_0^{(1)}
    ~~\text{with}~~
    a_0^{(1)} \sim \pi_1(a|s_0), ~~ a_0 \neq a_0^{(0)}
\end{align}
Upon convergence,\footnote{To judge convergence, humans can use a quantitative auxiliary measure to monitor performance. Since no feedback is based on this measure, it is not accompanied with the usual safety concerns. 
} $\pi_1$ produces length-1 sequences optimally, with respect to the parent's preferences.
The agent then matures to a new policy, $\pi_2$, initialised to $\pi_1$ and trained through feedback on recorded sequences of length $T=2$.\footnote{The increment $T=1,2,\ldots$ is appropriate for gridworld but may need modification in other contexts; see footnote \ref{foot:coarse_grain}.}
Exploitative length-2 recordings take the form
\begin{align}
    \Sigma^{(0)} &= s_0 \, a_0^{(0)} \, s_1^{(0)} \, a_1^{(0)} \nonumber\\[5pt]
    a_t^{(0)} &= \underset{a}{\text{argmax}}~\pi_{2-t}(a|s_t) ~~\text{for}~~ t = 0, 1
\label{eq:pi2_exploit}
\intertext{while exploratory sequences are drawn as}
    \Sigma^{(1)} &= s_0 \, a_0^{(1)} \, s_1^{(1)} \, a_1^{(1)} \nonumber\\[5pt]
    a_0^{(1)} &\sim \pi_2(a|s_0) ~~\text{with}~~ a_0 \neq a_0^{(0)} \nonumber\\[5pt]
    a_1^{(1)} &= \underset{a}{\text{argmax}}~\pi_1(a|s_1^{(1)})
\label{eq:pi2_explore}
\end{align}
The goal here is to optimise action choice for length-2 sequences.
Since the final state-action pair in each sequence is a length-1 sub-sequence, $\pi_1$ is already trained to draw this action optimally.
Thus, while length-2 recordings are drawn using both $\pi_2$ and $\pi_1$, they should be used solely to train $\pi_2$.
This is compatible with \Eq{loss} (where the $\pi$'s should have subscripts $T$ for completeness).

Once $\pi_2$ converges, the agent matures to $\pi_3$. Recordings of length $T=3$ are drawn from $\pi_3$, $\pi_2$, and $\pi_1$ analogously to \Eqs{pi2_exploit}{pi2_explore}. 
Through maturation, the agent's behaviour optimises for progressively longer sequences.

An example might clarify why recordings are drawn sequentially from \mbox{$\pi_T$, $\pi_{T-1}$, \ldots, $\pi_1$}. 
In chess, suppose the parent is only smart enough to see 1 move ahead, and that $\pi_1$ is already trained.
For $\pi_2$ to learn to see 2 moves ahead, the agent should present sequences $s_0\,a_0\,s_1\,a_1$ to the parent, where $a_0$ is chosen with $\pi_2$ but $a_1$ is not. Even if $\pi_2$ can detect a checkmate 2 moves from $s_1$, the human will not realise the value of the move and may penalise the sequence, because the human does not know the optimal state-action value function. Instead, $a_1$ should be chosen with $\pi_1$, which is already optimised with respect to the parent's preferences when there is one move to go.

Importantly, maturation only requires the parent to recognise improvements in the agent's performance; the human need not understand the agent's evolving strategy (see \Sec{exp_maturation}).

\section{Experiments}
\label{sec:results}

To test the safety of our \Parenting algorithm in a controlled way, we performed experiments in the AI Safety gridworlds of \citet{gridworlds}, designed specifically to capture the fundamental safety problems of \Sec{intro}.
Select gridworlds are shown in \Fig{gridworlds} and described in \Sec{safety_tests} below.

\subsection{Experimental Setup}

\subsubsection{Network Architecture}
\label{sec:architecture}

We used a neural-network policy $\pi(a|s)$ that maps the state of gridworld to a probability distribution over actions.
The state $s$ is represented by an $H \times W \times O$ matrix, where $H$ and $W$ are the gridworld's dimensions and $O$ is the number of object-types present; this third dimension gives a one-hot encoding of the object sitting in each cell.
There are 4 possible actions $a$ in any state: up, down, left, right.

The neural network has two components. 
The local component maps the local state $\ell(s)$, comprised of the agent's 4 neighbouring cells, through a dense layer of 64 hidden units, to an output layer with 4 linear units.
The global component passes the full state $s$ through several convolutional layers before mapping it, through a separate dense layer with 64 hidden units, to a separate output layer with 4 linear units.
All hidden units have rectifier activations. 
The convolutional processing includes up to 4 layers\footnote{The number of layers depends on the dimensions of the gridworld and are chosen to take the state matrix down to $2 \times 2$. 
} with kernel size $3 \times 3$, stride length 1, and filter counts 16, 32, 64, 64.
The local and global output layers are first averaged, then softmaxed, to give a probability distribution over actions.
This setup was implemented using Python 2.7.15 and TensorFlow 1.12.0.

\subsubsection{Hyperparameters}
\label{sec:algo_summary}

The \Parenting algorithm of \Sec{algo} has several hyperparameters.
Unless noted otherwise, we set
\begin{equation*}
    p_\text{guid} = 0.5 \qquad
    p_\text{rec} = 0.1 \qquad
    p_\text{pref} = 0.05 \qquad
    p_\text{train} = 1 
\end{equation*}
and held the recording length constant at $T=1$.
(Maturation is tested separately in \Sec{exp_maturation}.)
We also included entropy regularisation \cite{williams1991function} to control the rigidity of the agent's policy, with coefficients $\lambda_\text{global} = 0.01$ and $\lambda_\text{local} = 0.001$ for the separate neural-network policy components. We used Adam with default parameters for optimisation \cite{ADAM}.

\subsubsection{Substitute for Human Parent}
\label{sec:human_sub}

For convenience, we did not use an actual human parent in our experiments. 
Instead we programmed a parent to respond to queries in the following way.

We assume the parent has an implicit understanding of a reward function $r$ the agent should optimise and has intuition for a safe policy $\pi_p$ the agent could adopt. This is reasonable given the context assumed in \Sec{intro}. Furthermore, we assume the parent favours sequences \mbox{$\Sigma = s_0 \, a_0 \, \cdots \, s_{T-1} \, a_{T-1}$} with greater total advantage:
\begin{equation}
\label{eq:adv}
    \alpha(\Sigma) = \sum_{t=0}^{T-1} \left[ Q_p(s_t, a_t) - V_p(s_t) \right]
\end{equation}
where $V_p$ ($Q_p$) is the state (state-action) value function with respect to $\pi_p$ \cite{sutton1998introduction}.
In a deterministic environment, this quantity is equivalent to:\footnote{One could also impose a discount factor $\gamma$ on the sequence. We kept $\gamma=1$ except in \Secs{unsafe_exploration}{exp_maturation}, where we set $\gamma = 0.9$.}
\begin{equation}
\label{eq:adv2}
    r_0 + \cdots + r_{T-1} + V_p(s_T) - V_p(s_0)
\end{equation}
i.e.~the total reward the parent could accrue as a result of sequence $\Sigma$ (both during and after) minus what the parent expected to accrue following the baseline $\pi_p$ instead.

To motivate these assumptions, experiments in psychology suggest that human feedback does not correspond directly to a reward function \cite{ho2018people}. Instead, \citet{COACH} argue that humans do naturally base feedback on an advantage function. 
What is novel in our implementation is that the advantage is computed with respect to the parent's safe baseline policy -- without requiring an understanding of the agent's evolving policy.
Experiments with real human feedback in more complicated environments are needed to test whether this is reasonable in general.
Note also that since we compute \Eq{adv} exactly in gridworld, our experiments assume perfect human feedback. This assumption should be relaxed in more realistic future tests.

\subsubsection{Pre-Training}
\label{sec:pretraining}

Unless noted otherwise, our agent enters \Parenting after pre-training with policy gradients \cite{sutton1998introduction} to solve general path-connected mazes containing a single goal cell.
The reward function in these mazes grants $r=+50$ for reaching the goal and $r=-1$ for each passing time step. A pre-training step with Adam \cite{ADAM} was taken every 16 episodes, and \Parenting did not begin until the average reward earned per maze converged.
A pre-training step was also taken after each training step during \Parenting, to ensure this knowledge is not forgotten.

In general, pre-training reduces \Parenting's requisite human effort by allowing humans to focus on subtle safety concerns, rather than problems safely solved by other means.

\subsection{Safety Tests}
\label{sec:safety_tests}

Here we describe our experiments on the AI Safety problems of \Sec{intro}, highlighting the components of the \Parenting algorithm that solve each.

\subsubsection{Unsafe Exploration}
\label{sec:unsafe_exploration}

For unsafe exploration, we performed experiments in the gridworld of \Fig{safe_exploration}. Parental input was given according to \Eq{adv2} 
with a reward function that grants the light-blue agent $r=+1$ for reaching the green goal, $r=0$ for remaining on land, and $r=-1$ for falling in dark-blue water, which terminates the episode. 
We experimented with:
\begin{itemize}
    \item Traditional RL: used policy gradients as in pre-training 
    \item Direct Policy Learning: set $(p_\text{guid},
    \,p_\text{rec}, \,p_\text{pref}, \,p_\text{train})$ to $(0, \,0.5, \,0.1, \,1)$ to disable guidance queries
    \item Lax \Parenting: default hyperparameters  (\Sec{algo_summary}), $(p_\text{guid},
    \,p_\text{rec}, \,p_\text{pref}, \,p_\text{train}) =  
    (0.5, \,0.1, \,0.05, \,1)$
    \item Conservative \Parenting: cautious hyperparameters, $(p_\text{guid},
    \,p_\text{rec}, \,p_\text{pref}, \,p_\text{train}) =  
    (0.99, \,0.1, \,0.05, \,1)$
\end{itemize}

To emphasise the exploration required, we did not pre-train agents here.
We trained our agent from scratch to optimality 1000 times with each of the 4 algorithms and monitored the average number of water-deaths in each trial. The mean and standard deviation of the training deaths for each algorithm are shown in \Tab{safe_exploration}, along with the number of parenting queries used.
The agent suffered thousands of training deaths before reaching an optimal policy with traditional RL,\footnote{Similar experiments  in \cite{gridworlds} using modern RL algorithms yielded roughly comparable results.} compared to just 0 or 1 with conservative parenting.
This demonstrates the effectiveness of human guidance and direct policy learning at mitigating unsafe exploration.

\subsubsection{Reward Hacking}
\label{sec:reward_hacking}

Reward hacking is modelled in \Fig{reward_hacking}, 
where the blue agent must water dry yellow plants, which then turn green. Plants turn dry with 5\% probability per time step. The agent can ``reward hack'' by stepping in the turquoise bucket of water, which makes the entire garden appear watered and green. If the agent calculates rewards by counting green cells, it will be attracted to this dangerous policy.
\Parenting avoids this problem through its reliance on human input, as the parent will never prefer a clip in which the agent steps in water.
Since this environment is ideal to test maturation, we postpone experimental results to \Sec{exp_maturation}.

\begin{table}[!tb]
\caption{Deaths in the Unsafe Exploration gridworld before optimal policy was learnt. ``Lax'' and ``Conservative'' refer to different hyperparameter choices. Each table entry $\mu\pm\sigma$ was computed from 1000 trials.}
\label{tab:safe_exploration}
\vspace{-2mm}
\begin{center}\begin{small}
\renewcommand{\arraystretch}{2}
\begin{tabular}{>{\flushleft}m{1.4cm}>{\centering}m{1.6cm}>{\centering}m{1.6cm}>{\centering\arraybackslash}m{1.6cm}}
\toprule
& \shortstack[c]{\\Training\\Deaths} & \shortstack[c]{\\Guidance\\Queries} & \shortstack[c]{\\Preference\\Queries} \\
\midrule
\vspace{-1mm}
{Traditional\\RL} & 2300 $\pm$ 700 & -- & -- \\
\vspace{-1mm}
{Direct\\Policy\\Learning}  & 47 $\pm$ 14 & -- & 51 $\pm$ 13 \\
\vspace{-1mm}
{Lax\\\Parenting} & 15 $\pm$ 7 & 25 $\pm$ 5 & 6 $\pm$ 5 \\
\vspace{-1mm}
{Conservative\\\Parenting} & 0.6 $\pm$ 0.8 & 49 $\pm$ 8 & 0 \\
\bottomrule
\end{tabular}
\end{small}\end{center}
\vspace{-2mm}
\end{table}

\subsubsection{Negative Side Effects}
\label{sec:side_effects}

Negative side effects are addressed in \Fig{side_effects}, where the blue agent must navigate to the green goal in the presence of a movable dark-cyan box. Pushing the box into a corner is an irreversible action, representing a real-life irreparable side effect (e.g.~a broken vase). 
While going around the box or moving it reversibly is desired, the agent can reach the goal fastest by pushing the box down into the corner. 
If rewards are based solely on speed, the agent will adopt this dangerous behaviour. 
In contrast, since the parent would never reinforce a highly undesirable action, \Parenting is not susceptible to negative side effects.
This environment is also useful for testing whether behaviours learnt through \Parenting are generalisable or simply memorised; we thus postpone a discussion of results to \Sec{generalisability}.

\subsubsection{Unsafe Interruptibility}

Unsafe interruptibility is represented in \Fig{safe_interruptibility}, where the agent must navigate to the goal in the presence of a pink interruption cell and a purple button. 
If the agent enters the pink cell, there is a 50\% chance it will be frozen there for the remainder of the episode, prevented from reaching the goal. 
Upon pressing the purple button, the pink cell disappears along with the threat of interruption. 

If the agent simply gets rewarded for speed in reaching the goal, it will learn to press the button -- not a safely interruptible policy.
\Parenting, in contrast, is safely interruptible because the parent would never favour a clip of the agent avoiding human interruption, and there are no rewards left on the table if an episode is terminated early.

To test this, we parented an agent in \Fig{safe_interruptibility} for 50 queries, then checked whether its argmax policy involved pressing the purple button. In 100 repeated trials, it never did.

\subsubsection{Absent Supervisor}

The absent supervisor problem is modelled in \Fig{absent_supervisor}. Parental input is based on \Eq{adv2} 
where the reward function assigns \mbox{$r=+50$} for reaching the green goal and \mbox{$r=-1$} each passing time step. 
If the supervisor is present, represented by red side bars, there is a punishment of \mbox{$r=-10$} for taking the shortcut through the yellow cell.
With the supervisor absent, the punishment disappears.

\Parenting naturally gives no signal when the agent's actions are not viewed by a supervisor, so we parented\footnote{For this experiment, we used the default hyperparameters of \Sec{algo_summary} except for $\lambda_\text{local} = 1$ to weaken dependence on $\pi_\text{local}$.} our agent for 50 queries in the present-supervisor gridworld. 
Upon deployment in the absent-supervisor gridworld, we checked whether its argmax policy involved stepping in the yellow cell. 
In 100 repeated trials, it never did.

Because \Parenting omits feedback on unsupervised actions, the absent supervisor problem becomes an issue of distributional shift \cite{sugiyama2017dataset}. 
As long as the supervisor's absence does not cause an important change in the agent's environment, its policy should carry over intact. 
(To reiterate: in \Parenting, no signal is associated with the supervisor's ``leaving''.)
We tested this in gridworld as well: in $96 / 100$ trials reported above, the agent's policy remained optimal with supervisor removed
(while in $4 / 100$, the supervisor's absence ran the agent into a wall).

\subsection{Validation of Maturation}
\label{sec:exp_maturation}

Being the most complex of the gridworlds (with $2^{13}$ configurations of watered and dry plants) the Reward Hacking environment described in \Sec{reward_hacking} is ideal for testing maturation. 
For this experiment, the parent responds to queries as in \Sec{human_sub}, with a reward function that grants $r=+1$ for a legitimate plant-watering and $r=0$ otherwise.

Suppose the parent's policy $\pi_p$ is to water plants in a repeating clockwise trajectory around the garden's perimeter -- a good perpetual strategy, but suboptimal for short episodes. 
Nevertheless, the reliance of \Parenting on human judgement should not limit the agent's potential for optimisation.
When judging recordings of length $T=1$, the parent will prefer clips in which the agent successfully waters a dry plant, even if by an anti-clockwise step -- see \Eq{adv} or \eqref{eq:adv2}.
The agent will thus learn to go against $\pi_p$ and take a single anti-clockwise step, if it earns an extra watering. 
Upon maturation to $T=2$, the agent will learn to take 2 anti-clockwise steps, if it offers an advantage over $\pi_p$.
The agent will thus learn to outperform its parent.

\begin{figure}
\centering
\includegraphics[height=5cm]{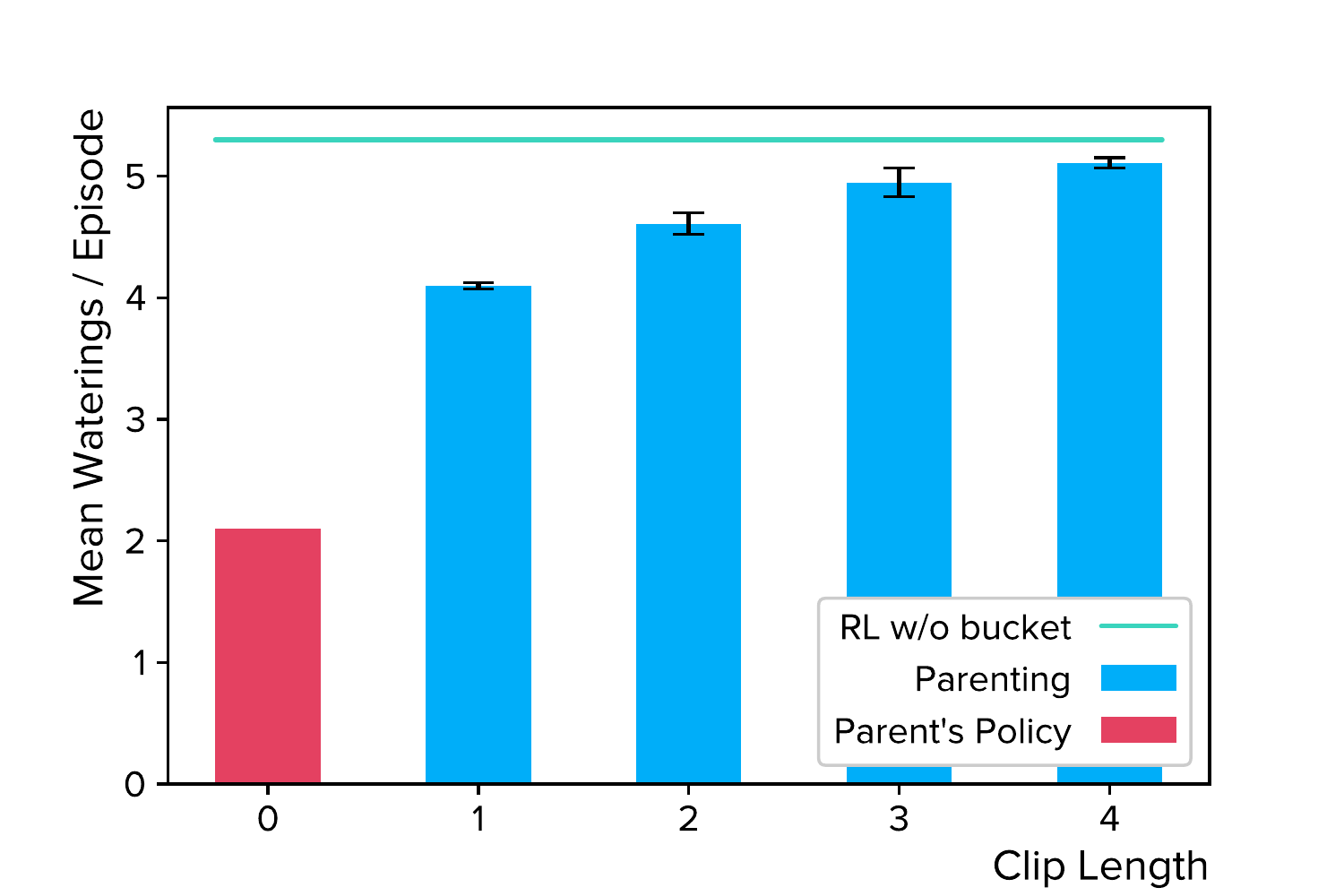}%
\caption{Maturation of agent's policy toward optimality in Reward Hacking environment. By learning from human feedback on lengthier recordings of its behaviour, the agent gradually optimises its policy to outperform its parent and approach the effectiveness of traditional RL.}%
\label{fig:maturation}%
\end{figure}

To test this, we set the episode length to 10 and initialised the gridworld to \Fig{reward_hacking}. We parented our agent for 1000 queries at each clip length $T$ before maturing to clips of length $T+1$, using $T = 1, 2, 3, 4$. \Fig{maturation} shows the resulting mean-waterings-per-episode at each stage, each mean being computed over 1000 episodes. The entire experiment was repeated 3 times to compute the standard deviations on the means (error bars in the figure). For comparison, policy gradients were used to train an RL agent to convergence (with the unsafe bucket cell removed from the environment!) whose mean score is also shown.
While the parent's policy $\pi_p$ achieves roughly 2 waterings per episode, the RL agent exceeds 5.
Despite this, maturation takes the agent to near-optimality, confirming its effectiveness.
\Parenting thus provides a safe avenue for autonomous learning agents to solve problems competently and creatively.
\vspace{2mm}

\subsection{Generalisability of Parented Behaviours}
\label{sec:generalisability}

It is important to understand whether \Parenting teaches behaviours abstractly, allowing lessons learnt to generalise, or if the agent merely memorises its parent's preferred trajectory.
Generalisability is critical for real-world applications.
Consider a manufacturer that parents household robots in a variety of environments, both simulated and real, so that customers would have little extra parenting required for customisation at home.
In this context, pre-training is analogous to first using RL to teach the robot to navigate rooms in safe simulations, to reduce the required parenting by the manufacturer's employees.

We tested generalisability in the Side Effects environment of \Sec{side_effects}.
To begin, we randomly generated path-connected gridworlds like \Fig{side_effects} that contain 1 goal, 1 box, any number of walls, and that are solvable only by moving the box. We discarded those generated gridworlds that the pre-trained agent could already solve. We kept 50 unique gridworlds satisfying these requirements, designating $n = 10$ of them for parenting and setting aside 40 for pre-parenting.

\begin{figure}
\centering
\includegraphics[height=5cm]{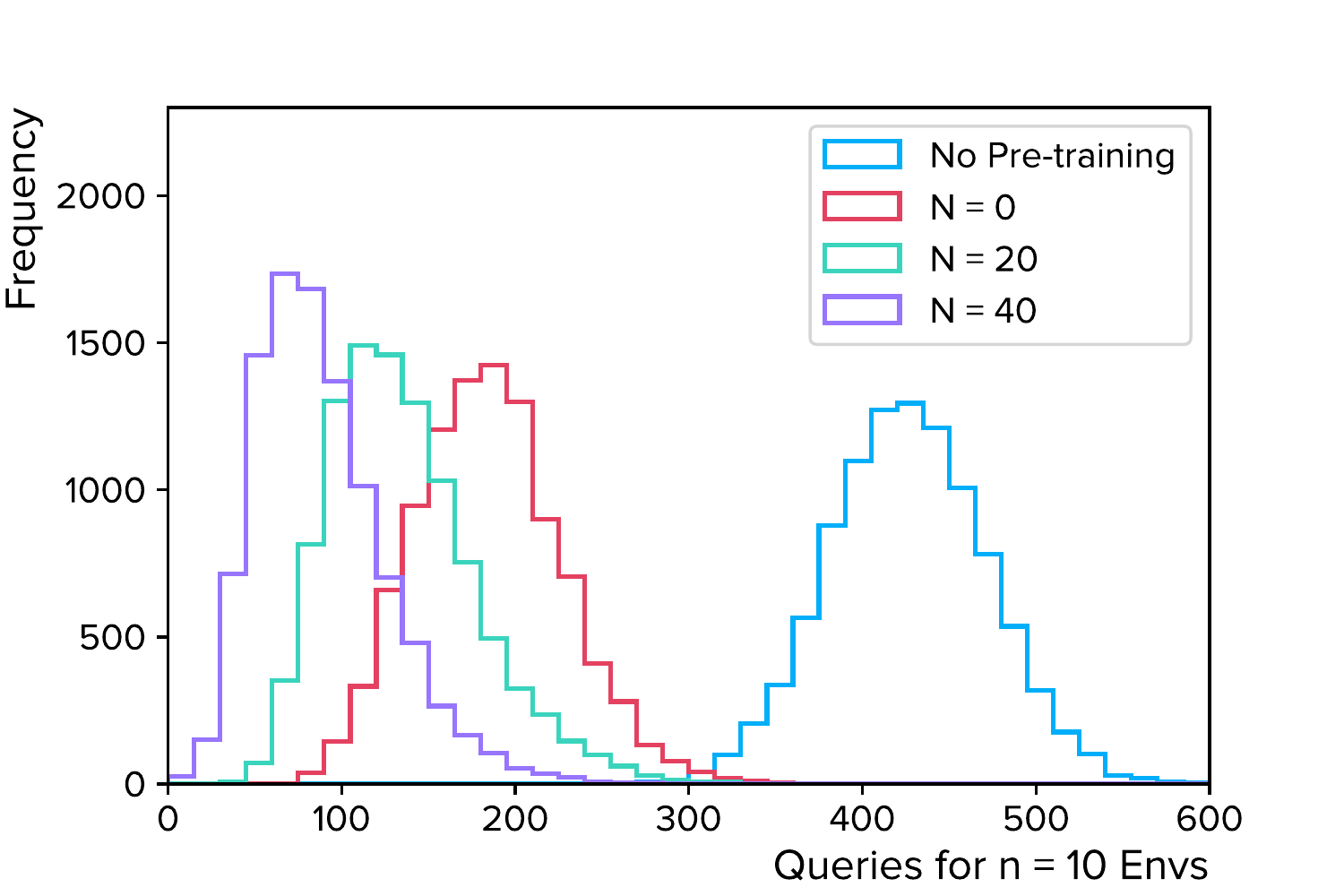}%
\caption{Generalisability of parenting to new environments. Agent was pre-trained to solve mazes then pre-parented to solve $N = 0$, $20$, or $40$ unique Side Effects environments, before being parented to optimality in $n=10$ held-out Side Effects environments. Queries required for held-out gridworlds are histogrammed -- 10,000 trials for each $N$.}%
\label{fig:generalisability}%
\end{figure}

For one experiment, we took an agent that was not pre-trained and parented it from scratch to optimality in the $n=10$ designated gridworlds (cycling through them during training). 
We repeated this for 10,000 trials and histogrammed the number of required queries in \Fig{generalisability}.
For the other experiments, we took a pre-trained agent and pre-parented it in \mbox{$N = \;$0, 20, or 40} of the set-aside environments before parenting in the $n=10$ designated gridworlds. 
The corresponding histograms in \Fig{generalisability}  show the benefits of pre-training and pre-parenting.
More pre-parenting reduces the number of queries required for safe operation in new environments, thus confirming \Parenting's generalisability.

\section{Discussion}
\label{sec:discussion}

In this section, we provide theoretical arguments that motivate our design of maturation and direct policy learning.

\subsection{Maturation as Value Iteration}
\label{sec:value_iteration}

The maturation process of \Sec{maturation} effectively optimises the agent's policy because of its connection to value iteration in dynamic programming \cite{sutton1998introduction}. 
To demonstrate this, we will make the same assumptions of \Sec{human_sub}. (These include perfect human feedback, which is necessary for this idealised discussion.) Let us also assume that $\pi_T$ converges in all relevant regions of state-space before maturation to $\pi_{T+1}$.

We will work in a deterministic environment for clarity, so that the parent's preferences on sequences of length $T$ are determined by computing\footnote{We omit $-V_p(s_0)$ since it drops out of comparisons when clips are chosen with the same (or sufficiently similar) initial states.}
\begin{equation}
    r_0 + \cdots + r_{T-1} + V_p(s_T)
\end{equation}
on each sequence $\Sigma = s_0 \, a_0 \, \cdots \, s_{T-1} \, a_{T-1}$,
with $V_p$ defined with respect to the parent's policy $\pi_p$.
Under these assumptions, maturation is equivalent to value iteration.
To show this, we prove maturation trains $\pi_T(a|s_0)$ to maximise $r_0 + V_{T-1}(s_1)$, where
\begin{equation}
    V_{T}(s_0) = \max_{a_0} \, \left[ r_0 + V_{T-1}(s_1) \right]
\end{equation}
for each $T = 2, 3, 4, \ldots$ with the base case
\begin{equation}
    V_1(s_0) = \max_{a_0} \, \left[ r_0 + V_p(s_1) \right]
\end{equation}
For $T=1$, $\pi_1(a|s_0)$ is trained to optimise $r_0 + V_p(s_1)$ because the parent responds to preference queries based on this quantity.
For $T=2$, sequences are recorded by successively drawing from $\pi_2(a|s_0)$ then $\pi_1(a|s_1)$, as prescribed in \Sec{maturation}. 
The involvement of $\pi_1$ implies that $\pi_2(a|s_0)$ is trained to maximise $r_0 + r_1 + V_p(s_2) = r_0 + V_1(s_1)$ as required.
Assuming the claim is true for sequence lengths through $T-1$, the argument for $T$ is similar: Because sequences are recorded by drawing actions successively from \mbox{$\pi_T$, $\pi_{T-1}$, \ldots, $\pi_1$} this implies that $\pi_T(a|s_0)$ is trained to maximise $r_0 + \cdots + r_{T-1} + V_p(s_T) = r_0 + V_{T-1}(s_1)$. 
The claim is thus proved by induction.

Note that $V_T$ is the same quantity that appears in value iteration \cite{sutton1998introduction}, which converges to optimality.
The agent's policy $\pi_T$ thus progressively outperforms the parent's policy $\pi_p$ as $T$ increases. Importantly, this process does not require the human parent to understand the agent's improving policy, just to recognise improving performance.

\subsection{Ambiguity in Preference-Based Reward Models}
\label{sec:ambiguity}

If, in contrast to \Parenting, one uses human preferences to learn a \emph{reward model}, there are subtleties one needs to overcome to ensure the corresponding \emph{optimised policy} is consistent with human desires.
We include this discussion here as it influenced the design of our algorithm.

\begin{figure}
{\hfill
\subfigure[Irreversible Side Effect]{
\includegraphics[height=3.1cm]{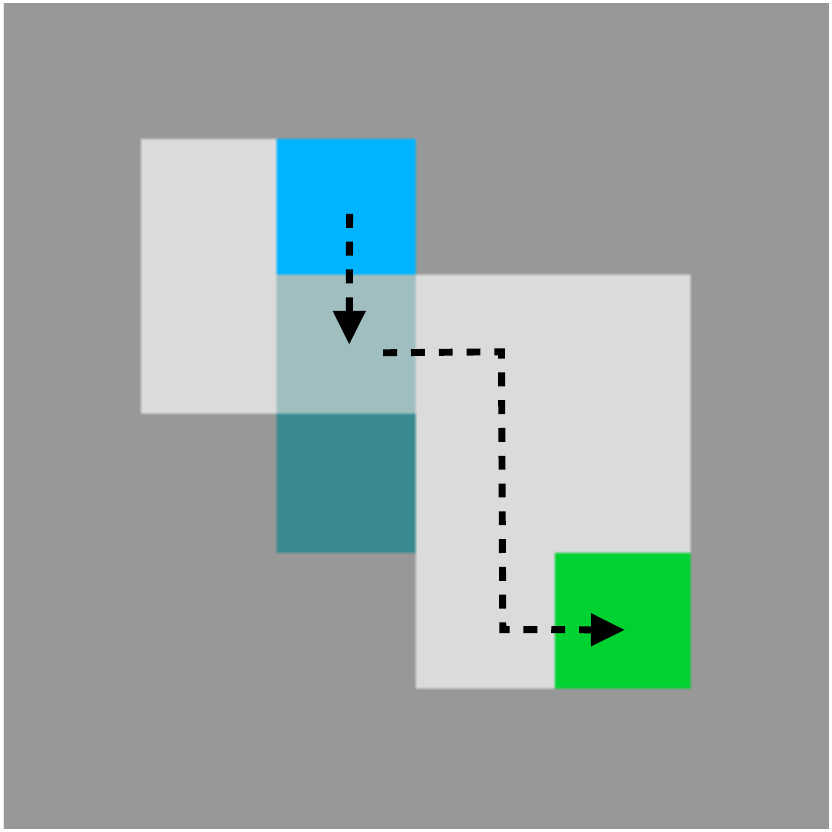}
\label{fig:side_effects_unsafe}}
~
\subfigure[Safe Solution]{
\includegraphics[height=3.1cm]{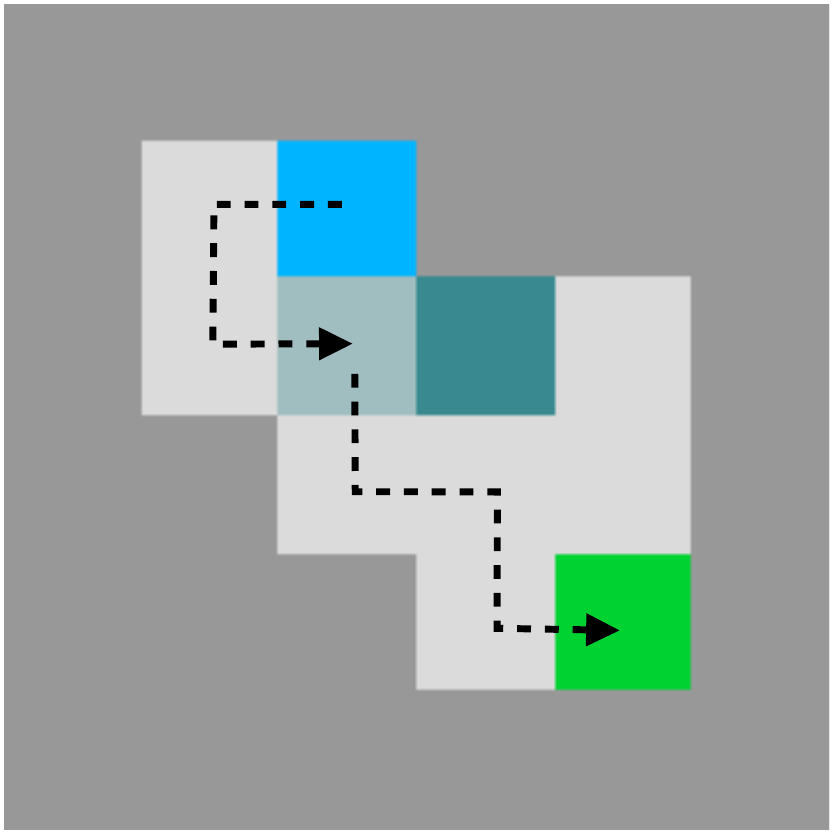}
\label{fig:side_effects_safe}}
\hfill}
\caption{Two trajectories in the Side Effects gridworld. Unsafe trajectory (a) optimises one reward function, $\rho$, while the safe solution (b) optimises a shifted function $\sigma = \rho - 1$.}
\label{fig:side_effects_detail}
\end{figure}

Suppose recordings $\Sigma = s_0 \, a_0 \, \cdots s_{T-1} \, a_{T-1}$ of fixed length are shown to a human in pairs to obtain preference data.
Let us assume that the human intuitively understands a reward function $r$ and (in this section only) favours clips $\Sigma$ that earn more reward $r(\Sigma) = \,r_0 + \cdots + r_{T-1}$. 
Then one could fit a reward model $\rho$ to the preference data. 
However, there is a shift ambiguity in the model, since both $\rho$ and $\sigma = \rho + a$ (for $a \in \mathbb R$) each describe the data equally well:
\begin{equation}
    \rho(\Sigma^{(0)}) - \rho(\Sigma^{(1)}) = \sigma(\Sigma^{(0)}) - \sigma(\Sigma^{(1)})
\end{equation}
This ambiguity can be eliminated by fixing the mean reward value. 
However, the reward function's mean can have a substantial effect on the learnt behaviour. 
See \Fig{side_effects_detail} for an example.
Suppose reward model $\rho$ grants $+1$ for reaching the goal, $-1$ for an irreversible side effect, and $0$ otherwise.
Then the trajectory of \Fig{side_effects_unsafe} accrues $\sum_t \rho_t = 0$, while the trajectory of \Fig{side_effects_safe} earns $\sum_t \rho_t = +1$. 
Optimisation of $\rho$ would thus avoid the irreversible side effect.
However, with a shifted reward model $\sigma = \rho - 1$, the unsafe trajectory scores $\sum_t \sigma_t = -5$, while the safe trajectory accumulates $\sum_t \sigma_t = -6$.
Optimisation of $\sigma$ would thus cause an irreversible side effect, against the human's wishes.

This problem can be overcome in practice by experimentally tuning the mean of the reward function, as well as its moments. However, this hyperparameter tuning would need to be repeated for each new task and would introduce a new type of unsafe exploration (of hyperparameter space).

This problem occurs because human preferences on same-length sequences are shift-invariant with respect to the reward function, while reinforcement learning is not. \Parenting avoids this problem through direct policy learning, which respects the symmetries of human preferences and thus does not require problem-by-problem tuning.\footnote{The hyperparameters of \Sec{algo_summary} control the rate of mistakes and speed of learning, rather than affecting the agent's learnt policy.}

\section{Conclusion}
\label{sec:conc}

In the context of near-future autonomous agents operating in environments where humans already understand the risks,
\Parenting offers an approach to RL that addresses a broad class of relevant AI Safety problems. We demonstrated this with controlled experiments in the purpose-built AI Safety gridworlds of \citet{gridworlds}. Importantly, the fact that \Parenting solves these problems is not particular to gridworld; it is due to the fact that humans can solve these problems, and \Parenting allows humans to safely teach RL agents. 
Furthermore, we have seen that two potential downsides of \Parenting can be overcome: 
(i) through the novel technique of maturation, a parented agent is not limited to the performance of its parent; and 
(ii) parented behaviours generalise to new environments, which can be used to reduce requisite human effort in the learning process. 
We hope the framework introduced here provides a useful step forward in the pursuit of a general and safe RL programme applicable for real-world systems.

\section*{Acknowledgements}

This work was developed and experiments were run on the Faculty Platform for machine learning.
The authors benefited from discussions with Owain Evans, Jan Leike, Smitha Milli, and Marc Warner. The authors are grateful to Jaan Tallinn for funding this project.

\bibliography{parenting_arxiv}
\bibliographystyle{icml2019}

\end{document}